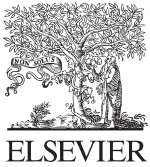

Available online at www.sciencedirect.com

ScienceDirect

International Conference on Advanced Computing Technologies and Applications (ICACTA-2015)

# Delaunay Triangulation on Skeleton of Flowers for Classification

Y H Sharath Kumar[a], N Vinay Kumar[b]*, D S Guru[c]

[a,b,c]*Department of Studies in Computer Science, Manasagangotri, University of Mysore, Mysore-570 006, India*

**Abstract**

In this work, we propose a Triangle based approach to classify flower images. Initially, flowers are segmented using whorl based region merging segmentation. Skeleton of a flower is obtained from the segmented flower using a skeleton pruning method. The Delaunay triangulation is obtained from the endpoints and junction points detected on the skeleton. The length and angle features are extracted from the obtained Delaunay triangles and then are aggregated to represent in the form of interval-valued type data. A suitable classifier has been explored for the purpose of classification. To corroborate the efficacy of the proposed method, an experiment is conducted on our own data set of 30 classes of flowers, containing 3000 samples.

*Keywords:* Flower Segmentation; Discrete Curve Evolution; Delaunay triangulation; Symbolic representation; Classification.

## 1. Introduction

Flowers and the ability to recognize them have been attracting humans for over a longer period of time. The taxonomy of flowers originally contained approximately 8000 plants, but since then it has been extended to encompass more than 250000 flower species around the world [1]. The flower images captured in a real time will create a several challenges like variations in viewpoint, scale, illumination, partial occlusions, and multiple instances. All these challenges need a very refined algorithm to classify flowers. Also, the cluttered background makes the problem more difficult, as we need to classify the flower image from the background. Moreover, the greatest challenge lies in preserving the intra-class and inter-class variabilities. The classification and grading of flowers will be the major part in floriculture. For formers, the floriculture has become one of the important commercial trades in agriculture owing to steady increase in demand of flowers.

* Corresponding author. Tel.: +91-9632454934.
  E-mail address: vinaykumar.natraj@gmail.com

In agriculture, floriculture is an area that deals with the trade, nursery, pot plants, seed & bulb production, micro propagation and extraction of essential oils from flowers. In these situations, flowers are identified and recognized manually. Even, the flower recognition process used for searching the patent flower images (whether the flower image applied for patent is already present in the patent image database or not? (Das et. al., [2]) involves manual interaction. Since these activities are being done manually and they are mainly labor dependent, automation is necessary.

We have gone through the survey of flower classification and found couple of interesting works that has published in this direction. Firstly, Nilsback and Zisserman [15] designed a flower classification system by extracting visual vocabularies which represent color, shape and texture features of flower images. In [6], Guru et al. developed a neural network based flower classification using different combinations of texture models such as color texture models [18], gray level co-occurrence matrix, gabor responses. Guru et al., [7] designed a flower classification system using only whorl region of flower and also compared with entire region of flower. The features like gray level co-occurrence matrix, gabor responses are used. The features are fed into classifier for classification. Saitoh et al., [5] describe an automatic recognition system for wild flowers using both flower and leaf from each image.

For segmentation of flowers from the background, the concept of interactive graph cuts are used in [4]. In [17], the authors have proposed a two step model for segmentation of flowers; initially they separate the foreground & background regions and then they extract the petals structure from the segmented flower.

For representation of flowers, [15] have used shift invariant feature transform (SIFT) descriptors which preserves the shape features; and also used MR8 filter bank responses which preserve the textural features of flowers. Also the authors use the combination of all the three (color, shape and texture) visual vocabularies with different weights in order to study the effectiveness of the different features. For experimentation, [15] considered a dataset consists of 17 flower classes, each with 80 flower images and they recorded an accuracy of 71.76% for all three combinational features. Nilsback and Zisserman in their work [16] considered a large data set of 103 classes; each consists of 40 to 250 samples in order to study the efficacy of classification system. They have used the low level features such as color, histogram of gradient orientations and also SIFT features to get an accuracy of 72.8% using (multiple) SVM classifier.

In the literature, we can also find couple of works done on indexing and quantitative evaluation of flower classification system. Das et al., [2] proposed an indexing method to index flower patent images using the domain knowledge. In their work the image color is mapped to names using ISCC-NBS color system and X Window system. Then, each flower image is discretized in HSV color space and each point on the discretized HSV space is mapped to a color name in ISCC-NBS and X Window system in order to index the flowers. Yoshioka et al., [3] performed a quantitative evaluation on flower image where they consider a petal colors for computing principal components. The quantitative evaluation indicates that the different PCs correspond to different color features of petals such as color depth, and difference in color depth of upper & lower parts of an image.

Apart from texture, colour, shape and their fusion features used for classifying the flowers, there are very few attempts found in the literature towards classification of flowers based on their internal structure, a skeleton. In [13], authors have extracted skeletons from flower images using skeleton pruning method and they have also preserved the shape context feature from the skeleton. For classification, they have used NN classifier. In [14], authors have extracted the DCT features from the junction points and end points of the flower skeleton and preserved the same in the form of an interval valued type features. For classification, they have used symbolic classifier. It shall be interesting to note that in [14] there is an improvement in results compared to [13] because of interval valued representation of features. So we feel that if spatial topological relationships among the endpoints and junction points can be preserved in addition to interval valued representation, we may have much better results. Indeed the same thing has been revealed through our experimentation.

The organization of the paper is as follows: section 3 presents the proposed methodology which includes flower segmentation, Skeleton Pruning, Delaunay Triangulation of skeleton points and symbolic representation. In section 4 dataset and experimental results obtained using the proposed model are presented. The paper is concluded in section 5.

## 2. Proposed Methodology

The proposed method has five stages: Segmentation, Skeleton extraction, Delaunay triangulation, Representation and Classification. The flowers are segmented using whorl based region merging segmentation and then flower skeleton is obtained using skeleton pruning method. The junction points and end points on skeleton are used to generate the triangles using Delaunay triangulation and are represented through symbolic approach. Finally the symbolic classifier is designed to classify the flowers.

*2.1 Segmentation*

The flowers are segmented using whorl based region merging segmentation proposed in [14]. In flowers, whorl region is indentified using Gabor filter response and marked as foreground; and boundary is marked as background. The flower is initially divided into regions using quick shift segmentation and later regions are merged into foreground & background using maximal region merging. Detailed discussion of segmentation is presented in [9].

*2.2 Skeleton Pruning*

After segmenting a flower from background region, we then extract a skeleton of a flower for effective representation. For generating the skeleton of a flower, we have adopted the work proposed by Bai et.al, [10]. Initially, we binarize the gray scale flower image and then by applying some morphological operations, we will get the skeleton of a flower image. Then, the Discrete Curve Evolution (DCE) is used to simplify the polygons obtained from un-pruned flower skeleton image. It recursively removes the least relevant polygon vertices, where the relevance measure is computed with respect to the actual partially simplified versions of the polygon. Then the skeleton is pruned so that only branches ending at the convex DCE vertices will remain. The pruned skeleton is guaranteed to preserve the topological relationship of the shape and it is robust to noise and boundary deformation. The main benefit of using DCE is the fact that DCE is context sensitive.

After obtaining a skeleton from a flower, we extract some points from skeleton viz., end points and junction points that are useful in generating Delaunay triangulation. A skeleton point having only one adjacent point is called an endpoint (the skeleton endpoint); a skeleton point having three or more adjacent points is called a junction point. The figure 1 shows the detected endpoints and junction points for a given flower skeleton. In the subsequent section we can see how these points are helpful in generating Delaunay triangulation.

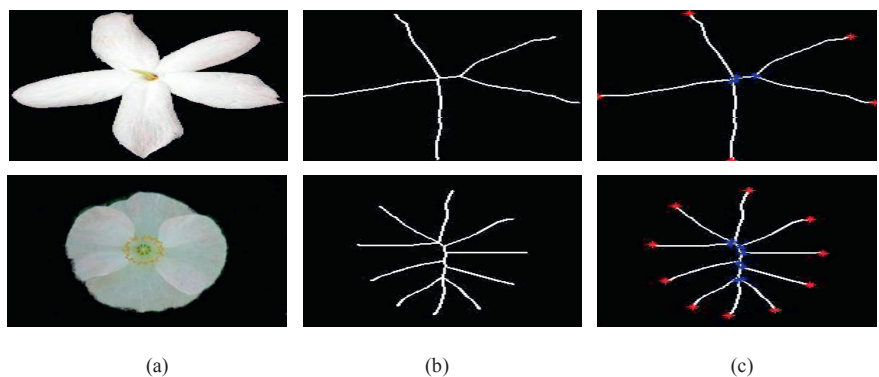

(a)                              (b)                              (c)

Fig. 1.  a) Flower image, b) Skeleton of the respective flower image, c) End points and junction points marked on skeleton image

*2.3 Triangular Representation of Skeleton*

In this section, we present a method to classify a flower using Delaunay Triangles. For the skeleton flower we find the endpoints and junction points. Using the detected points we generate the triangles. Later for each triangle we extract features like length and orientation which are represented by interval data. For the purpose of classification we design a voting based classifier.

*2.3.1  Delaunay Triangulation*

Triangulation is a process of dividing a region/space into sub regions/sub spaces in the form of triangles. The space may be of any dimension, however, a 2D space is considered here, since we are dealing with 2D points (junction and end points). Our goal is to associate a 2 dimensional topological structure from the skeleton points (junction and end). To associate such topological structure from skeleton points, we have used a triangulation method called Delaunay Triangulation [12].

Consider a set *P* of 2 dimensional points $p_1, p_2, \ldots, p_z$, we can compute the Delaunay triangulation of *P* by first computing its Voronoi diagram. The Voronoi diagram decomposes the 2D space into regions around each point $p_i$, such that all the points in the region around $p_i$ are closer to it compare to any other points in *P*. Given the Voronoi diagram, the Delaunay triangulation can be formed by connecting the centers of every pair of neighbouring Voronoi regions [12]. So, the Delaunay triangulation is very desirable in our application, since the computation of the geometric transformations among flowers is based on corresponding skeleton triangles.

*2.3.2  Flower Triangulation*

The proposed flower classification system represents flowers in terms of their skeletons. The flower skeleton contains a set of identified junctions and end points which are represented by coordinates $(x, y)$. Using identified coordinates $(x, y)$, the Delaunay triangulation is computed. Figure 2(b) demonstrates the Delaunay triangulation for the flower skeleton shown in figure 2(a).

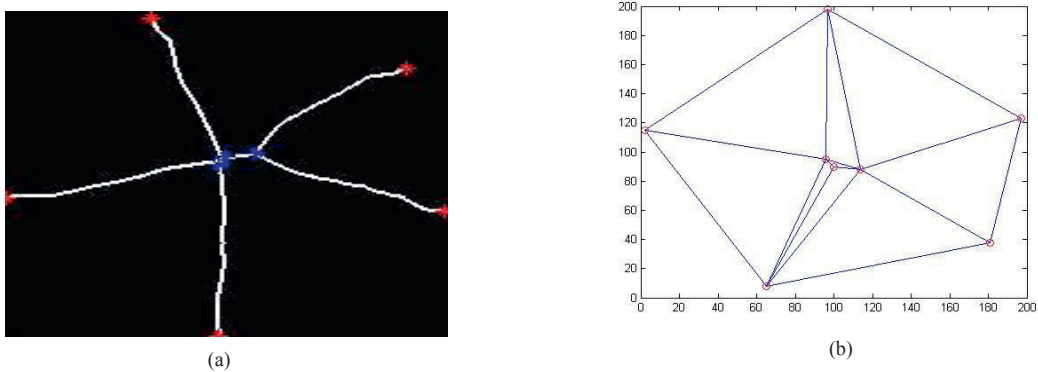

(a)        (b)

Fig. 2. a) Skeleton of the flower image, b) Triangles obtained after applying Delaunay Triangulation to (a)

**3. Feature Extraction**

If we apply Delaunay Triangulation for P, a set of 'z' points (junction points and end points), the number of triangles that we obtain is (2z-2-k), where, k is the number of points on the convex hull of P [12]. Therefore, for each flower sample we get 'm' triangles, where m= (2z-2-k).

After obtaining 'm' triangles for a given skeleton image, we have extracted features like length and angle of each side of a triangle that are invariant to geometrical transformations like translation, rotation and flipping.

The lengths of all sides of each triangle are obtained using Euclidean distance measure and to determine angles we have used Cosine Rule (Eqn. 1), Sine Rule (Eqn. 2), and sum of all the angles of a triangle rule (Eqn. 3).

$$A = \cos^{-1}\left(\frac{a^2 + b^2 - c^2}{2*a*b}\right) \quad (1)$$

$$C = \sin^{-1}\left(\frac{a*\sin(A)}{c}\right) \quad (2)$$

$$B = (180 - A - C) \quad (3)$$

Here a, b, and c are the lengths of each sides of a triangle obtained by computing the distance between the coordinates of the Delaunay triangles and A, B and C are the opposite internal angles of the sides a, b, and c respectively.

After computing the lengths and angles of all three sides of each triangle, we have totally 6m (mx6 matrix) number of features representing a single flower sample. Hence, there are 6mn (n x m x 6) numbers of flower samples representing a class $C_j$ as shown in figure 3(a).

Further, we reduce the extracted features for symbolic representation of flower samples. As flower skeletons have a considerable intra class variation in each subgroup, using conventional data representation, preserving these variations is difficult. Hence, the proposed work is intend to use unconventional data analysis called symbolic data analysis which has an ability to preserve the variations among the data more effectively [8].

In this work, we have assimilated the lengths and angles of all samples belong to a class $C_j$ in the form of an interval-valued type feature, where mx6 crisp features are converted into 1x6 interval valued type features. The assimilation is done by considering the minimum lengths, maximum lengths and minimum angles, maximum angles due to all m triangles of the classes. In the same way, we have computed inter-valued features for all samples of all classes and then the computed feature matrix is stored in the database as shown in Figure 3. Now the size of the feature matrix is (Nxn)x6, consisting 6 inter-valued type features for each Nxn sample.

Formally, let $[S_1, S_2, S_3, \ldots, S_n]$ be a set of *n* samples of a flower class say $C_j$; $j = 1, 2, 3, \ldots, N$ (N denotes number of classes) and let $F_i = [(l_{i11}, l_{i12}, l_{i13}), (l_{i21}, l_{i22}, l_{i23}), (l_{i31}, l_{i32}, l_{i33}), \ldots, (l_{im1}, l_{im2}, l_{im3})]$ and $Th_i = [(\alpha_{i11}, \alpha_{i12}, \alpha_{i13}), (\alpha_{i21}, \alpha_{i22}, \alpha_{i23}), (\alpha_{i31}, \alpha_{i32}, \alpha_{i33}), \ldots, (\alpha_{im1}, \alpha_{im2}, \alpha_{im3})]$ be the set of features (length & angle) obtained from *m* number of triangles in a flower sample $S_i$ of the class $C_j$.

Let $(MinL_{i1}, MinL_{i2}, MinL_{i3})$ and $(MaxL_{i1}, MaxL_{i2}, MaxL_{i3})$ be the minimum and maximum (length) feature values respectively, obtained from all the *m* triangles of $S_i$.

$MinL_{ip} = \min(l_{ikp})$, $MaxL_{ip} = \max(l_{ikp})$

Where k=1, 2, …, m; i=1,2,…,n and p=1,2,3.

Similarly, let $(MinA_{i1}, MinA_{i2}, MinA_{i3})$ and $(MaxA_{i1}, MaxA_{i2}, MaxA_{i3})$ be the minimum and maximum (angle) feature values respectively, obtained from all the *m* triangles of $S_i$.

$MinA_{ip} = \min(\alpha_{ikp})$, $MaxA_{ip} = \max(\alpha_{ikp})$

Where k=1, 2, …, m; i=1,2,…,n and p=1,2,3.

Now, we recommend capturing the different variations of $i^{th}$ sample in the form of interval valued feature. That is, each sample $S_i$ is represented by the use of interval valued features

$$([l_{ip}^-, l_{ip}^+]) \text{ and } ([\alpha_{ip}^-, \alpha_{ip}^+]) \quad (4)$$

Where i=1,2,3, …, n and p=1,2,3.

where $l_{ip}^- = MinL_{ip}$ and $l_{ip}^+ = MaxL_{ip}$ and $\alpha_{ip}^- = MinA_{ip}$ and $\alpha_{ip}^+ = MaxA_{ip}$

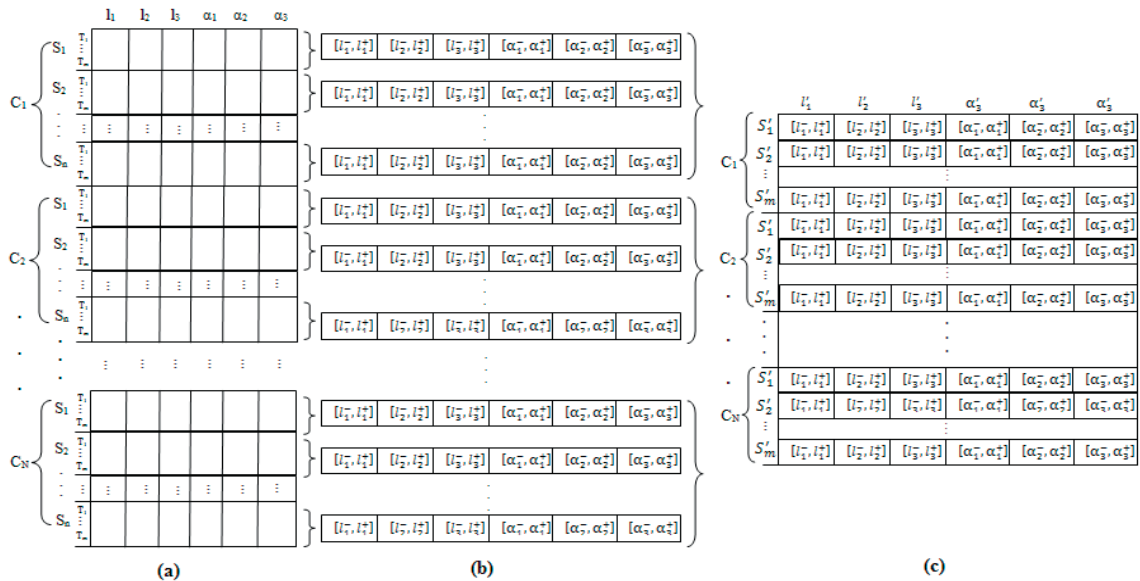

Fig. 3. Feature Matrix Representation a) Conventional Feature Matrix representation, (b) Assimilated length and angles Features (c) Symbolic Feature Matrix

Each interval representation depends on the minimum and maximum of respective individual features. The interval $[l_{ip}^-, l_{ip}^+]$ and $[\alpha_{ip}^-, \alpha_{ip}^+]$ represent the upper and lower limits of the length and angle feature values of the sample $S_i$. Now, a reference flower sample of a class $C_j$ is formed by the use of interval type and is given by

$$RF_{ji} = \{([l_{ip}^-, l_{ip}^+]),([\alpha_{ip}^-, \alpha_{ip}^+])\} \quad (5)$$

where $i = 1, 2,...,n$ ; $j = 1, 2,...,N$ and p=1,2,3. It shall be noted that unlike conventional feature vector, this is a vector of interval-valued features and this symbolic feature vector is stored in the knowledgebase as a representative of the $i^{th}$ sample of $j^{th}$ class. We recommend computing symbolic feature vectors for each individual sample of a class and store them in the knowledgebase for future recognition requirements. Thus, the knowledgebase has $n*N$ number of symbolic vectors of dimension 6.

## 4. Classification

In this section we adopt the symbolic classifier (Guru and Prakash, [11]) with a slight modification for classifying the flowers. During classification, a test sample of an unknown flower is described by a set of *m* features (length, orientation) of type crisp and compares it with the corresponding interval type features (length, orientation) of the respective symbolic reference samples $RF_{ji}$ stored in the knowledgebase to ascertain the efficiency.

Let $FL_t = [(l_{t11},l_{t12},l_{t13}),(l_{t21},l_{t22},l_{t23}),(l_{t31},l_{t32},l_{t33}),...,(l_{tm1},l_{tm2},l_{tm3})]$ and $FA_t = [(\alpha_{t11},\alpha_{t12},\alpha_{t13}),(\alpha_{t21},\alpha_{t22},\alpha_{t23}),(\alpha_{t31},\alpha_{t32},\alpha_{t33}),...,(\alpha_{tm1},\alpha_{tm2},\alpha_{tm3})]$ be m x 3 dimensional vectors representing a test flower. Let $RF_{ce}$; $c = 1,2,3,...,N$, $e=1,2,...,n$ be the representative symbolic feature vectors stored in knowledgebase. During flower classification process each $k^{th}$ (where k=1,2,3) side/angle's (length and orientation of $q^{th}$ triangle, q=1,2,...,m) feature of a test flower is compared with the respective intervals of all the representatives of $k^{th}$ (where k=1,2,3) side/angle of $q^{th}$ triangle, (q=1,2,...,m) of $e^{th}$ sample in $c^{th}$ class, to examine whether the feature value of the test flower lies within them. Such features are counted as acceptance count. The test flower sample is said to belong to the class $C_j$ with which it has a maximum acceptance count $AC_j$.

Acceptance count $AC_j$ for $j^{th}$ class is given by,

$$AC_j = ACL_j + ACA_j \quad (6)$$

$$ACL_j = \sum_{k=1}^{m} \sum_{p=1}^{3} \sum_{i=1}^{n} C(l_{tkp}, [l_{ip}^-, l_{ip}^+]) \tag{7}$$

$$ACA_j = \sum_{k=1}^{m} \sum_{p=1}^{3} \sum_{i=1}^{n} C(\alpha_{tkp}, [\alpha_{ip}^-, \alpha_{ip}^+]) \tag{8}$$

Where,　j=1,2,3, ..., N,

$$C(l_{tkp}, [l_{ip}^-, l_{ip}^+]) = \begin{cases} 1 & \text{if } (l_{tkp} \geq l_{ip}^- \text{ and } l_{tkp} \leq l_{ip}^+) \\ 0 & \text{otherwise} \end{cases}$$

$$C(\alpha_{tkp}, [\alpha_{ip}^-, \alpha_{ip}^+]) = \begin{cases} 1 & \text{if } (\alpha_{tkp} \geq \alpha_{ip}^- \text{ and } \alpha_{tkp} \leq \alpha_{ip}^+) \\ 0 & \text{otherwise} \end{cases}$$

From the above discussion, we can observe that how an unknown test sample is being classified as a known sample with a class label. Here, it can be achieved by taking into care of how many triangles from a test sample are matched against the triangles of a class each time, where the maximum triangles matched class is considered as a class for an unknown test sample. Finally, we can call this classification as voting based classification as the decision of a classifier is judged based on the maximum triangles matching with the respective class.

*4.1 Datasets*

In this work, we have created our own flower image database despite of existence of other databases, as these are less intra class variations or no change in view point. We have collected flower images from different sources like World Wide Web and in addition to it we have taken up some photographs of flowers that can be found in and around our place. We have a dataset consisting of 3000 images divided into 30 flower classes. Figure 5 shows the samples of randomly selected flowers from our database. It is clearly understood that there is a large intra class variation and a small inter-class variation make this dataset very challenging.

*4.2 Results*

In this experimentation we intend to study the accuracy of proposed symbolic feature in classifying a flower using voting based classifier. The experimentation has been conducted on database of 30 classes under varying number of training samples 40%, 60% and 80% from each class. We picked images randomly from the database and experimentation is conducted five times for each training percentage. Then we have taken the average of 5 trials and plotted the maximum, minimum and average classification accuracies for 40%, 60% and 80% of training samples respectively (Figure 4).

On 40% training samples the averaged (5 trials) classification accuracy achieves maximum of 82.26%, minimum of 77.73%, and average of 79.99%. On 60% training samples the averaged (5 trials) classification accuracy achieves maximum of 88.20%, minimum of 85%, and average of 86.66%. On 80% training samples the averaged (5 trials) classification accuracy achieves maximum of 94%, minimum of 92%, and average of 93%.

From the figure 4, we can observe that the classification accuracy of the proposed method increases with increase in the number of training samples. A comparative analysis is performed on various methods ([13] and [14]) with the proposed method and it has been tabulated in table 1. It shows that the proposed method yields better results compared to the other two methods for varied training samples.

*Y.H. Sharath Kumar et al. / Procedia Computer Science 45 (2015) 226 – 235* 233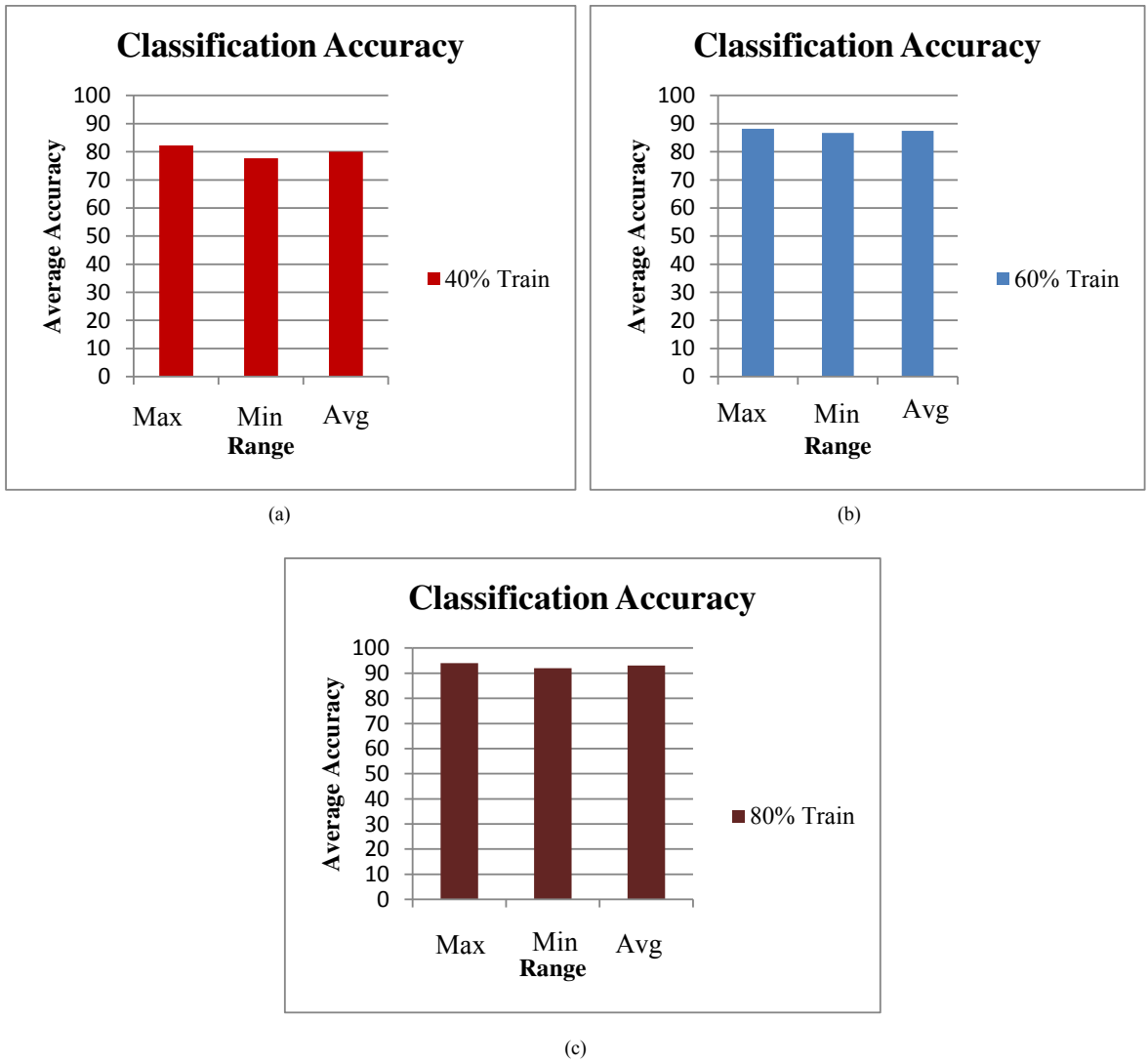

Fig. 4. Average classification accuracies obtained for 5 trials (a) 40% Training (b) 60 % Training (c) 80% Training

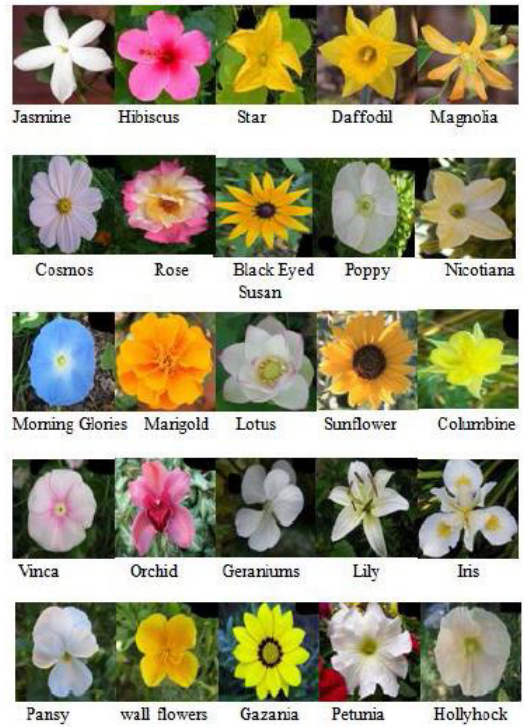

Fig. 5. Samples of flower images used in our experimentation

Table 1. Comparative analysis based on classification accuracies obtained for different training sets

| % of Training | Out of 5 trials | Method used in [13] using Nearest Neighbor Classifier | Method used in [14] using Symbolic classifier | Proposed Method using Voting based Classifier |
|---|---|---|---|---|
| 80 | Maximum | 87.1 | 90 | **94** |
| | Minimum | 85.88 | 87 | **92** |
| | Average | 86.49 | 89 | **93** |
| 60 | Maximum | 82.16 | 88.5 | 88.20 |
| | Minimum | 81.71 | 84.5 | **86.66** |
| | Average | 81.93 | 86.08 | **87.43** |

## 5. Conclusion

In this paper, we made a successful attempt to explore the applicability of the concepts of symbolic data to Triangle based flower classification. We explored the spatial topological relationships exist among junction points and endpoints. The newly proposed model has an ability to capture the variations of the features in training samples of flowers. The symbolic representation of the features reduces the time taken to classify a given test sample of an unknown flower. Also, we have modified the existing symbolic classifier into voting based classifier which is used for classification. In order to investigate the effectiveness and robustness of the proposed method, we have conducted extensive experiments on our own dataset and compared the results with other methods. In future we are planning to consider only partial occluded flowers for classification.

## Acknowledgements

The work by N. Vinay Kumar was financially supported by DST-INSPIRE Fellowship under Department of Science and Technology, Government of India, India.

## References

1. Linneaus C. Systemae Naturae. Impensis Direct. Laurentii Salvii, 1759.
2. Das M., Manmatha R., Riseman EM. Indexing flower patent images using domain knowledge. In IEEE Intelligent systems; 1999, vol. 14, p. 24 -33.
3. Yoshioka Y, Iwata H, Ohsawa R and Ninomiya S. Quantitative evaluation of flower color pattern by image analysis and principal component analysis of Primula sieboldii. E. Morren. Euphytica; 2004, p. 179 – 186.
4. Boykov Y, Jolly MP. Interactive graph cuts for optimal boundary and region segmentation of objects in N-D images. In: Proceedings of International Conference on Computer Vision (ICCV - 01); 2001, vol. 2, p. 105—112.
5. Saitoh, T, Aoki, K, Kaneko T. Automatic recognition of blooming flowers. In the Proceedings of 17th International Conference on Pattern Recognition; 2004, vol. 1, pp. 27-30.
6. Guru DS, Sharath YH., Manjunath S. Textural features in flower classification. Mathematical and Computer Modeling; 2011, 54(3-4), p. 1030-1036.
7. Guru DS, Sharath YH, Manjunath S. Classification of Flowers based on Whorl Region. 5th Indian International Conference on Artificial Intelligence; 2011, p. 1070-1088.
8. Billard L and Diday E. Symbolic data analysis: Conceptual statistics and data mining. Wiley series in computational statistics, 2006.
9. Guru DS, Sharath Kumar YH, Manjunath S. Whorl identification in flower: a Gabor based approach. Proceedings of the First International Conference on Intelligent Interactive Technologies and Multimedia; 2010, p. 172-178.
10. Bai L, Latecki J, Liu WY. Skeleton pruning by contour partitioning with discrete curve evolution. IEEE Trans. Patt. Anal. Mach. Intell; 2007, 29(3), p. 449–462.
11. Guru DS, Prakash HN. On-line signature verification and recognition: An approach based on symbolic representation. IEEE Transaction on Pattern Analysis and Machine Intelligence; 2009, 31, p. 1059- 1073.
12. Bradford Barber C, David Dobkin P, and Hannu Huhdanpaa. The Quickhull Algorithm for Convex Hulls. ACM Transactions on Mathematical Software; 1996, Vol. 22, No. 4, p. 469–483.
13. Guru DS and Sharath Kumar YH. Skeleton Based Approach for Flower Classification. International Journal of Machine Intelligence; 2011, Vol. 3, Issue 4, p.191-198.
14. Sharath Kumar YH and Guru DS. Classification of Flowers: A Symbolic Approach. International Conference on Multimedia Processing and Information Technology; 2013, ACEEE.
15. Nilsback ME and Zisserman, A. A Visual Vocabulary for Flower Classification. Proceedings of the IEEE Conference on Computer Vision and Pattern Recognition; 2006, vol. 2, p. 1447-1454.
16. Nilsback ME and Zisserman, A. Automated flower classification over a large number of classes. Proceedings of the Indian Conference on Computer Vision, Graphics and Image Processing; 2008, p.722-729.
17. Nilsback ME and Zisserman A. Delving into the whorl of flower segmentation. In the Proceedings of British Machine Vision Conference; 2004, vol. 1, p. 27-30.
18. Guru DS, Sharath Kumar YH, Manjunath S. Texture Features and KNN in Classification of Flower Images. IJCA, Special Issue on RTIPPR (1); 2010, p. 21-29.